\documentclass[letterpaper, 10 pt, conference]{ieeeconf}  
\IEEEoverridecommandlockouts                              

\usepackage{graphicx} 
\usepackage{color,soul}
\usepackage{hyperref}
\usepackage{mwe}
\usepackage{subcaption}
\usepackage{amsmath}
\usepackage{tabularx} 
\usepackage{multirow}

\newcolumntype{L}{>{\centering\arraybackslash}m{3cm}}
\usepackage{float}
\usepackage[thinc]{esdiff}
\usepackage{graphicx} 
\usepackage[table]{xcolor}
\usepackage{booktabs}
\usepackage{makecell}
\usepackage{cite}
\usepackage{textcomp}
\usepackage{caption}
\usepackage{microtype}
\renewcommand{\baselinestretch}{0.97}

\title{\LARGE \bf
HASHI: Highly Adaptable Seafood Handling Instrument for Manipulation in Industrial Settings
}
\author{Austin Allison$^{1*\dag}$, Nathaniel Hanson$^{1*}$, Sebastian Wicke$^{1}$, Taşkın Padır$^{1}$
\thanks{$^{*}$Equal contribution.}%
\thanks{This research is supported by the National Science Foundation under Award
Number 1928654.}
\thanks{$^{1}$Institute for Experiential Robotics, Northeastern University, Boston, Massachusetts, USA.}%
\thanks{\dag Corresponding author: \tt\small allison.a@northeastern.edu}%
\thanks{Project Code, Materials List, and Assembly Instructions: \url{https://river-lab.github.io/hashi/}
}
}
\begin{document}

\maketitle
\thispagestyle{empty}
\pagestyle{empty}

\begin{abstract}
The seafood processing industry provides fertile ground for robotics to impact the future-of-work from multiple perspectives including productivity, worker safety, and quality of work life. The robotics research challenge is the realization of flexible and reliable manipulation of soft, deformable, slippery, spiky and scaly objects. In this paper, we propose a novel robot end effector, called HASHI, that employs chopstick-like appendages for precise and dexterous manipulation. This gripper is capable of in-hand manipulation by rotating its two constituent sticks relative to each other and offers control of objects in all three axes of rotation by imitating human use of chopsticks. HASHI delicately positions and orients food through embedded 6-axis force-torque sensors. We derive and validate the kinematic model for HASHI, as well as demonstrate grip force and torque readings from the sensorization of each chopstick. We also evaluate the versatility of HASHI through grasping trials of a variety of real and simulated food items with varying geometry, weight, and firmness.

\end{abstract}

\section{Introduction}
\label{sec:intro}
In 2022, the United States imported record amounts of seafood, corresponding to a trade deficit of more than \$20 billion \cite{gephart2019create}. This is primarily due to lack of capacity for processing and reduced worker availability. Robotics and automation has the potential to impact the future of work in the seafood processing industry. To achieve this, there is a need to introduce novel designs, tools and methods and productivity measures. 

Deformable object manipulation is a challenging task in robotics. The industrial settings in seafood processing facilities pose additional challenges to maintain the integrity of the products while maximizing the yield. Through discovery interviews with more than 40 small-to-medium size enterprises (SMEs) in the seafood industry, located in the U.S., seafood hubs such as New Bedford, MA, and Kodiak Island, AK, we identified inspection and grading of delicate seafood items as a use-case to motivate the development of a novel robot end effector. More specifically, (1) to inspect seafood in a production line both visually and physically for quality and grading purposes, (2) to collect, sort and tray seafood for packaging by human workers, and (3) to orient, align, and flip fresh seafood on a production line. 

Our existing inventory of end effectors that can perform manipulation of deformable objects often features multiple articulated joints in the fingers \cite{andrychowicz2020learning}, or employs active surfaces to achieve the desired levels of control \cite{Cai2023}. However, these approaches increase mechanical complexity and are difficult to adopt in food processing where food safety is paramount. Soft grippers are also used to handle food products \cite{wang2017prestressed}; however, most widely-used designs are underactuated without the ability to reposition items in-hand.  With this framing in mind, the core question of this research is: \textit{Can a rigid robot end effector be both gentle, dexterous, and mechanically simple?}

\begin{figure}[t!]
    \centering
    \includegraphics[width=0.98\columnwidth]{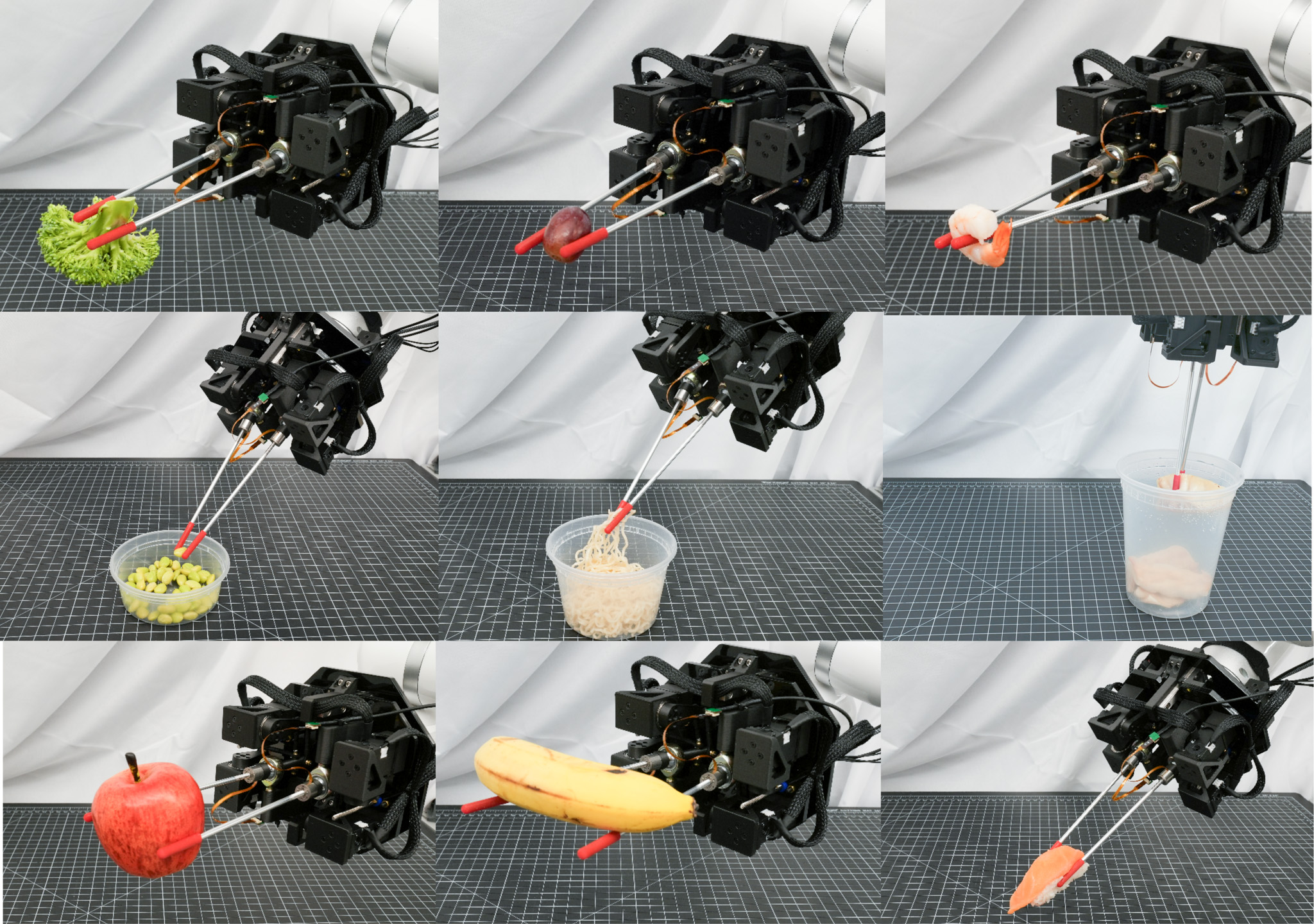}
    \caption{HASHI end effector mounted to 7 degrees of freedom (DOF) co-robot, manipulating a variety of deformable or fragile food items. The gripper also exhibits a series of grasp strategies including open-air pinch, in-container grasp, and large object lift.}
    \label{fig:HASHI_first_page}
    \vspace{-0.5em}
\end{figure}

Inspired by how chopsticks have become a versatile tool for handling food items among numerous cultures, we present a dexterous manipulator specifically targeted at handling food items. This research focuses on the development and validation of the hardware platform to enable chopstick-based motion planning for control and handling of objects. The specific contributions are:
\begin{itemize}
    \item Development of open-source hardware and software for a chopstick end effector for deformable object manipulation.
    \item Kinematic modeling for control of the mechanism and experimental verification of the models.
    \item Demonstration of a dual-chopstick gripper for the manipulation of food items of various size, hardness, and weight.
\end{itemize} 
\section{Related Work}
\label{sec:related}

\subsection{Food Processing and Manipulation}
Successful manipulation of food during processing is a difficult problem in robotics. This is attributed to the wide range of food items in size, shape, friction, and homogeneity \cite{Ishikawa2022}. Cooking food in any way also further changes the physical properties, size, shape, density, mechanical properties, coefficient of friction, and elastic modulus \cite{Williams2005, kadowaki2013}. Previous efforts have explored the use of grippers with tactile sensors \cite{Ishikawa2022} where delicate foods are manipulated with a simple parallel plate gripper, since it is the most common robot manipulator. Food manipulation is of particular importance to the assistive robotics community, where multiple efforts have been made to increase the sensorization of end effectors for feeding.\cite{Bhattacharjee2019,park2016towards}. Both of these papers illustrate the potential for haptic feedback in the transportation of food by an end effector.

\subsection{Food Gripper Designs}

 Although mechanically-simple end effectors, such as the parallel-plate gripper, have their place, many application-driven robots utilize a specialized end effector design. \cite{Wang2022} introduces a food gripper taxonomy by which a gripper is classified by its contact location with the food. Food grippers most similar to a parallel plate or three-pronged grippers grasp the food at the sides, including compliant grippers \cite{Qiu2023,Low2022}. \cite{ZWang2021} demonstrated an example of this formulation, where the jaws of a parallel plate gripper are replaced by rubber strings that conform to the sides of food items. In one configuration, this design can fully close its jaws, allowing it to lift a large range of food items. However, certain granular foods, such as beans or peas, are small enough to slip through the gaps between bands. Many unconventional grasping strategies are used to grasp from an angle other than top-down\cite{Amend2016}. These include the quad-spatula gripper \cite{Gafer2020} that scoops up a wide variety of compliant and fragile foods by sliding underneath, with the added benefit of a large planar workspace. Grippers that exploit additional forces such as the Bernoulli effect, \cite{Petterson2010}, also fall into this category. When piercing the food item is an acceptable condition, another class of grippers has been successfully demonstrated \cite{Wang2021,Endo2016}, albeit only when irreversible deformation does not degrade the quality of the food item. These grippers are also effective in grabbing numerous small food items (noodles or slices of green onion) simultaneously.

\subsection{Chopstick Related Work}

Previous designs have explored the potential for a chopstick-like end effector in both mechanical design and simulated capabilities. \cite{Yang2022} employed Bayesian Optimization and Deep Reinforcement Learning to predict different gripping and manipulation poses with chopsticks in-hand given a manipulation task.  \cite{Marcosticks2021} combined motion planning and stick positioning in-hand from a variety of pick and place actions; thus mirroring the natural variety of chopstick use found in multiple human users. \cite{Ke2020} explored the use of chopsticks for teleoperation with a single degree-of-freedom (DOF) chopstick to grasp unstable items such as a smooth sphere and a potato chip. They then leveraged this data for use in model-free imitation learning designed to combat the covariate shift \cite{Ke2021}.

There are only a handful of examples of robots designed with chopstick-like mechanisms for manipulation, and even fewer that are related to the food industry. Most recently, Dextrous Robotics has released the DX-1, a package unloading robot using two chopstick-like arms on rails that allow for linear motion in two directions and rotational motion about their mounts \cite{Dextrous}. They demonstrate manipulation of objects as small as a sugar cube and as large as an armchair. Conversely,\cite{Tanikawa1999} focused on dexterous manipulation of objects on the order of microns. \cite{Yabugaki2013} designed a teleoperated mechanism that has force feedback on the order of nano-Newtons. While this mechanism also has three DOF per chopstick, it was designed for surgical teleoperation with an effective range of motion that is far too minuscule to handle food other than individual granular media like salt or other spices. \cite{Sakurai2016} developed a surgical robot that replaces thick forceps with thin-diameter rods for laparoscopic surgery. \cite{Tadano2010} developed a system capable of positioning both tips of the chopstick at an $XYZ$ position in the workspace, as well as linear motion and rolling motion about the axial length of the chopstick, allowing the rotation of an object at the tips in two directions \cite{Tadano2010}. This dexterity allows for minimally invasive manipulation of surgical instruments as well as body tissue inside the abdominal cavity. 

\cite{koshizaki2010control,yamazaki2012autonomous,yamazaki2012various} demonstrated the use of a one-DOF chopstick mechanism in a series of works. Their design opens and closes one chopstick using a cam and is mounted to a frame with additional DOF for positioning and orienting the gripper; however, their system is meant to be placed on a tabletop, rather than on the end effector of a robot. Nevertheless, it is capable of many autonomous motions, such as cutting, mixing, stacking, piercing, and spooning a mass of small discrete items like rice. Their design employs a force-torque sensor in line with the stationary chopstick, as well as a computer vision system and a user interface capable of being operated by an individual with limb deficiencies. Our work builds upon the advances of the literature by providing a multi-DOF mechanism, small and light enough to be mounted as an end effector on existing manipulators, with kinematic modeling to support the manipulation of a wide variety of foodstuffs.

\section{Design}
\label{sec:design}

\begin{figure*}[t!]
   \vspace{0.5em}
   \centering
    \includegraphics[width=0.98\linewidth]{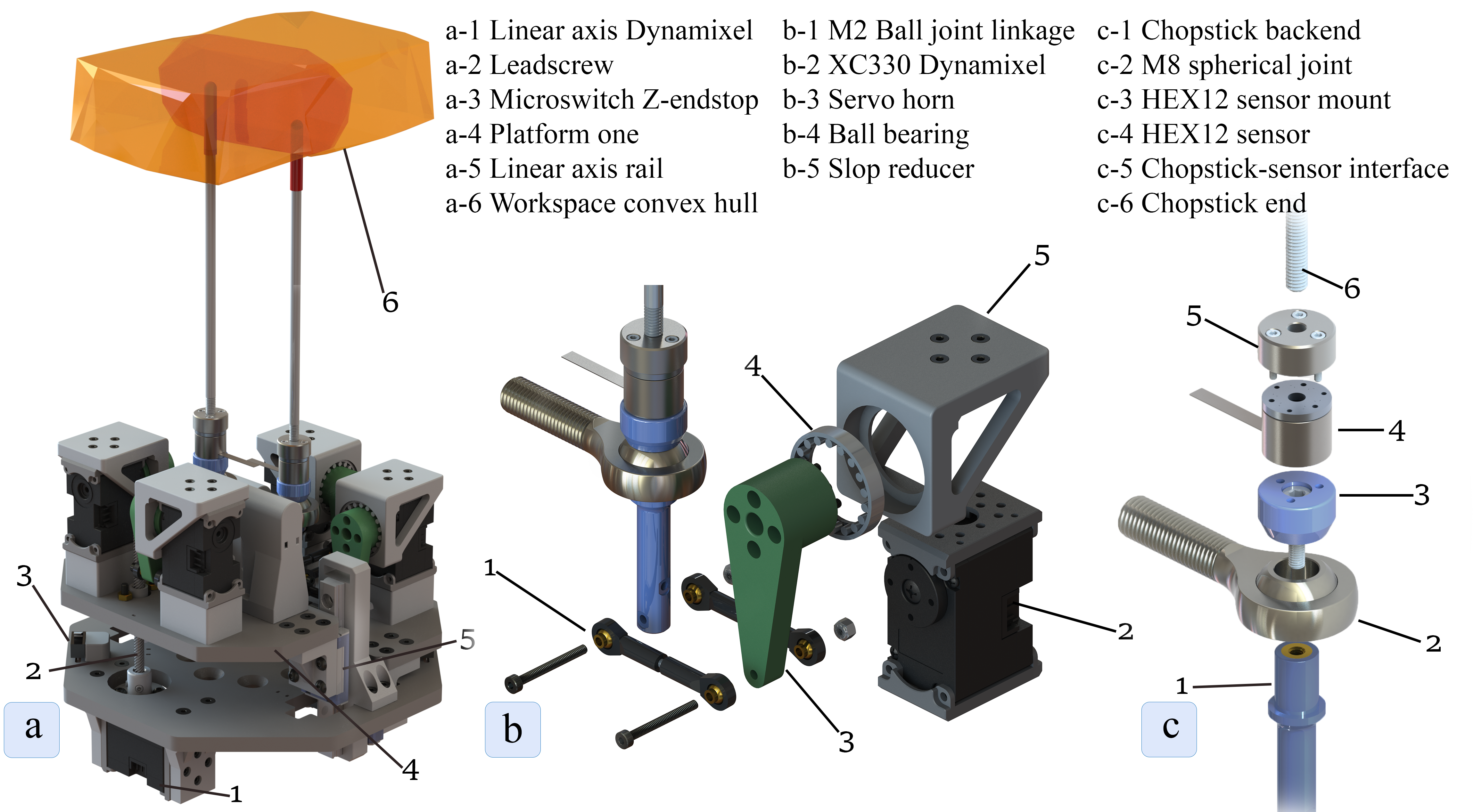}
    \caption{a) Sample image of the convex hull generated from the kinematic analysis overlaid in a CAD model to demonstrate individual and shared workspace of each chopstick b) Exploded diagram of yaw servo actuation with corresponding linkage and connection to multibody chopstick backend c) Exploded diagram of the multibody chopstick}
    \label{fig:hashi_mechanism}
     \vspace{-1.5em}
\end{figure*}

HASHI consists of two platforms capable of independent movement. Each platform houses one chopstick and has a linear travel range of 35 mm along the Z-axis. The height of the platform is controlled by a leadscrew mounted directly to the faceplate of a servo. To achieve successful manipulation in the desired range ($\pm50 \text{mm}$ in the $XY$ plane), Dynamixels from the XC330 line were selected for their torque output, size, ease of mounting, and compatibility with ROS \cite{quigley2009ros}. XC330-T288 servos were chosen for pitch and yaw actuation, and XC330-T181 servos were chosen for the linear axes. While HASHI is capable of piercing food items, the speed and torque limits for the servo were the primary design factors. These servos were mounted to the linear plate so that their axis of rotation was in the same plane as the pivot that the multibody chopstick assembly is mounted to. The intrinsic slop each servo has leads to imprecise motions. To combat this source of error, the circular body of each servo horn is enclosed with a ball bearing held in place by a slop reducer as seen in Fig. \ref{fig:hashi_mechanism}b. This reduces all slop except for the backlash inside the servo gearbox. The linear axis servo was not fitted with a slop reducer. This allows for some axial misalignment with the leadscrew and threaded flange and reduces binding under load. 

Each chopstick is instrumented with a 6-axis force/torque (F/T) sensor developed by Resense\footnote{\url{https://www.resense.io}}. Its size is only marginally larger than the diameter of the chopstick itself; moreover, it has an allowable force limit of $\pm25$ N and torque limit of $\pm125$ mNm. We selected this device for its mass ($\approx10$ g), measurement resolution (10 bit), and sampling rate (1 kHz). The addition of these sensors does not affect the range of motion nor force output. The location of these sensors at the top of the mechanism, and thoroughly out of contact with any crumbs or liquid stemming from the manipulation of food, increases the mechanical simplicity of the design, while simultaneously making the device food safe.

Each chopstick is an aluminum M4 partially threaded connecting rod. As the chopstick is directly connected to the F/T sensor, aluminum was chosen for its light weight, which minimizes sensor drift induced by gravity. The ability to quickly unscrew each chopstick increases ease of sterilization and decreases cross-contamination when working with a variety of foods. The tip of each chopstick is fitted with a custom silicone sock cast in Dragon Skin FX-Pro\texttrademark  (Smooth-On), which has a shore hardness of A-2. This addition increases friction while adding minimal mass to the tip. This sock is easily removed to reveal cut threads underneath, which are helpful to get purchase on objects that are slippery and deformable yet durable. All 3D printed parts were printed out of Markforged Onyx CF-Nylon, and all of the flat plates were lasercut out of 1/4" acrylic.

\subsection{Inverse Kinematics}
\begin{table}[b]
\caption{Specifications of HASHI Manipulator}
\label{tab:hashi_specificaitons}
\centering
\footnotesize
\setlength\tabcolsep{4 pt} 
\begin{tabularx}{0.9\columnwidth}{cp{0.5\columnwidth}c}
\toprule
    \textbf{Symbol} & \textbf{Description} & \textbf{Value}\\
    \midrule
    $l_c$ & Length of chopstick [mm] & $162$ \\
    $l_j$ & Length of M2 ball-joint linkage [mm] & $32.5$ \\
    $l_p$ & Length of pitch servo horn [mm] & $28$ \\
    $l_y$ & Length of yaw servo horn [mm] & $32$ \\
    \bottomrule
    \end{tabularx}
\end{table}

The multi-servo control mechanism does not fit within many pre-defined frameworks for forward or inverse kinematics. The Denavit-Hartenberg parameter \cite{denavit1955kinematic} formulation does not apply here due to the nonserial nature of the mechanism. Therefore, a more direct method must be used to derive the inverse kinematics. Formally, the inverse kinematic problem is defined as follows: Given an $XYZ$ position of the tip in the achievable workspace, we solve for the pitch and yaw servo angle, as well as the rotational position of the linear axis servo. 

\begin{figure}[t!]
    \includegraphics[width=\columnwidth]{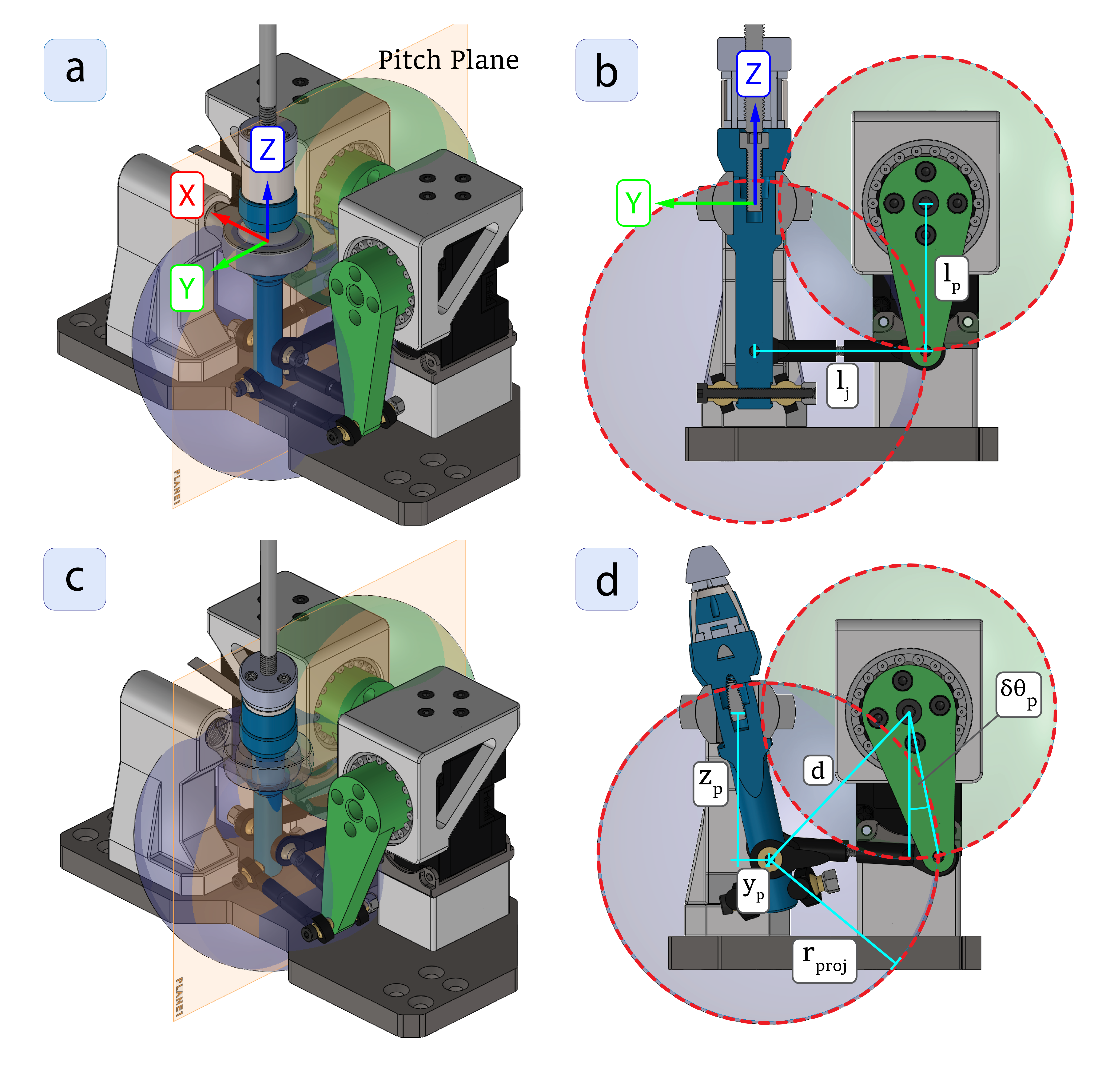}
    \caption{a) Isometric view of one platform with spherical interaction visualization and reference frame assignment b) Cross-sectional view of pitch servo actuation and circular intersections c) Isometric view showing components moved out of plane d) Repeated view of pitch servo showing new circular intersection point and servo position}
    \label{fig:spherical_cross_section}
    \vspace{-1.5em}
\end{figure}

We start by looking at one platform and modeling the motion of the multibody chopstick as a pivot about the M8 spherical ball joint. We assign a reference frame to this joint, shown in \ref{fig:spherical_cross_section}a. This allows us to exploit spherical coordinates by treating the chopstick as the R-vector \cite{Wolfram}, with a length of $l_c$, as described in Table \ref{tab:hashi_specificaitons}. To account for the specified linear travel of the platform, we must solve for $\Phi$ and $\Psi$ independently of the global Z given. This is done with the equation set below. First, $\Psi$ is calculated with Eqn.\ref{eqn:psi}, using the input X and Y coordinates. Using The Pythagorean Theorem in Eqn. ~\ref{eqn:z_calc} yields the theoretical $Z$-value in spherical coordinates of the chopstick, where $r$ is the projection of the R-vector in the $XY$ plane calculated with $\Psi$. From this we can now calculate $\Phi$ as in Eqn.~\ref{eqn:phi}. $z_{calc}$ allows us to calculate the vertical displacement of the entire platform in Eqn.~\ref{eqn:d_p}.

\begin{equation}
    \label{eqn:psi}
    \Psi = tan^{-1}\frac{X}{Y}
\end{equation}
\begin{equation}
    \label{eqn:z_calc}
    z_{calc} = \sqrt{l_c^{2}-r^{2}}
\end{equation}
\begin{equation}
    \label{eqn:phi}
    \Phi = cos^{-1}\frac{z_{calc}}{l_c}
\end{equation}
\begin{equation}
    \label{eqn:d_p}
    d_p = Z-z_{calc}-z_{offset}
\end{equation}
Now that we have the two spherical angles, representing the rotation of the chopstick base about the pivot, and its $Z$-position, we derive the relationship to the servo horn rotations for the pitch and yaw servos. Fig.~\ref{fig:spherical_cross_section} visualizes the important constructs of the subsequent derivation. Considering that this relationship maps a spherical motion to a linear motion about two axes, a linear relation between servo angle and chopstick position does not exist. Simply put, rotating the servo horns causes the chopstick to trace a spherical arc in 3D space, resulting in a loss of height proportional to the displacement from the $Z$-axis. We start by considering the interaction between a single servo and the multi-body chopstick backend via the M2 ball-joint linkage. We define two spheres, one centered on the M2 linkage mount on the multi-body chopstick backend, with radius $l_j$, and the other with the origin at the servo pivot with radius $l_s$. As the motion of the chopstick is constrained to the pivot and the motion of the servo arm is constrained to rotation in a single plane, we define a plane from this sub-assembly along the center of the servo horn to generate two circles from these spheres. The bottom intersection point of this circle is the desired location of the servo horn, which we solve for to obtain the desired servo angle. 

The location of the two spheres is found by the following method. By utilizing the defined frame at the chopstick pivot, we again exploit spherical coordinates to calculate the position of the sphere with origin on the chopstick backend:
\begin{gather*}
x_{pos}=-l_bsin(\Phi)sin(\Psi) \\
y_{pos}=-l_bsin(\Phi)cos(\Psi) \\
z_{pos}=-l_bcos(\Phi)
\end{gather*}
where $l_b$ is the distance from the reference frame to the connection point of either the pitch or yaw M2 linkage, and can be switched with $l_p$ or $l_y$ as shown in Fig. \ref{fig:spherical_cross_section}b. 

When considering the pitch servo, motion is constrained to the $YZ$-plane, and the origin of the second sphere is coincident with the axis of rotation of the pitch servo horn, with a radius of $l_p$ as well, as shown in Fig. \ref{fig:spherical_cross_section}b. When the chopstick backend moves around, the sphere tied to it will move out of the plane of interest, meaning that the circular cross section will change in radius and location as seen in Fig. \ref{fig:spherical_cross_section}c-d. Combining $x=0$ as the equation of the pitch plane with the general equation of the sphere describes the circular intersection of the two, and yields the following:

\begin{equation}
    (y-y_p)^{2}+(z-z_p)^{2}=\sqrt{r^{2}-x_p^{2}}
\end{equation}
The right hand side is the projected radius $r_{proj}$ and $x_p$ is the subtracted value for $a$ from the equation of a sphere. $b$ and $c$ are $y_p$ and $z_p$, respectively, from the equation set above, or more generally, the two coordinates defining the plane of interest.

Finally, we calculate the intersection points of the two circles. To do this, we find the distance between the origins, which is given by:
\begin{equation}
\label{eqn:circ_dist}
    d=\sqrt{(h_2-h_1)^{2}+(v_2-v_1)^{2}}
\end{equation}
Where $(h,y)$ correspond to the horizontal and vertical coordinates of the circle centers. The horizontal distance between the center of one circle and the radical line is the line that joins the two intersection points. This quantity is found by combining the two equations of the circles and solving for the horizontal coordinate of the two intersection points.
\begin{equation}
\label{eqn:horiz_intersection}
    l=\frac{r_1^{2}-r_2^{2}+d^{2}}{2d}
\end{equation}
 The vertical coordinate is then given by:
\begin{equation}
\label{eqn:vert_intersection}
    h=\pm\sqrt{r_1^{2}-l^{2}}
\end{equation}
Eqns. \ref{eqn:circ_dist},\ref{eqn:horiz_intersection},\ref{eqn:vert_intersection} consider the case where the two circular centers lie on an axis drawn between the two, and so a new axis between them must be defined when considering the origin at the M8 ball joint. We can then substitute the values for $d$, $l$, and $h$ into these general equations for horizontal and vertical positions while considering the pitch case in the $YZ$ plane. 
\begin{gather}
    y=\frac{l}{d}(y_2-y_1)\pm\frac{h}{d}(z_2-z_1)+y_1\\
    z=\frac{l}{d}(z_2-z_1)\pm\frac{h}{d}(y_2-y_1)+z_1
\end{gather}

After filtering out the intersection point outside the max ROM of the servo horn, the equations below are employed to determine the displacement angle of each servo, where $(y,z)$ and $(x,z)$ are the calculated intersection points in the pitch and yaw planes respectively, and $(y_{ps},z_{ps})$ and $(x_{ys},z_{ys})$ are the locations of the pitch and servo circles respectively.
\begin{gather}
    \delta_p=tan^{-1}\frac{y-y_{ps}}{z-z_{ps}}\\
    \delta_y=tan^{-1}\frac{x-x_{ys}}{z-z_{ys}}
\end{gather}

While only the calculation in the $YZ$ plane was shown, both calculations need to be done for every chopstick position.

\subsection{Kinematic Validation}
\begin{figure}[t!]
    \includegraphics[width=\columnwidth]{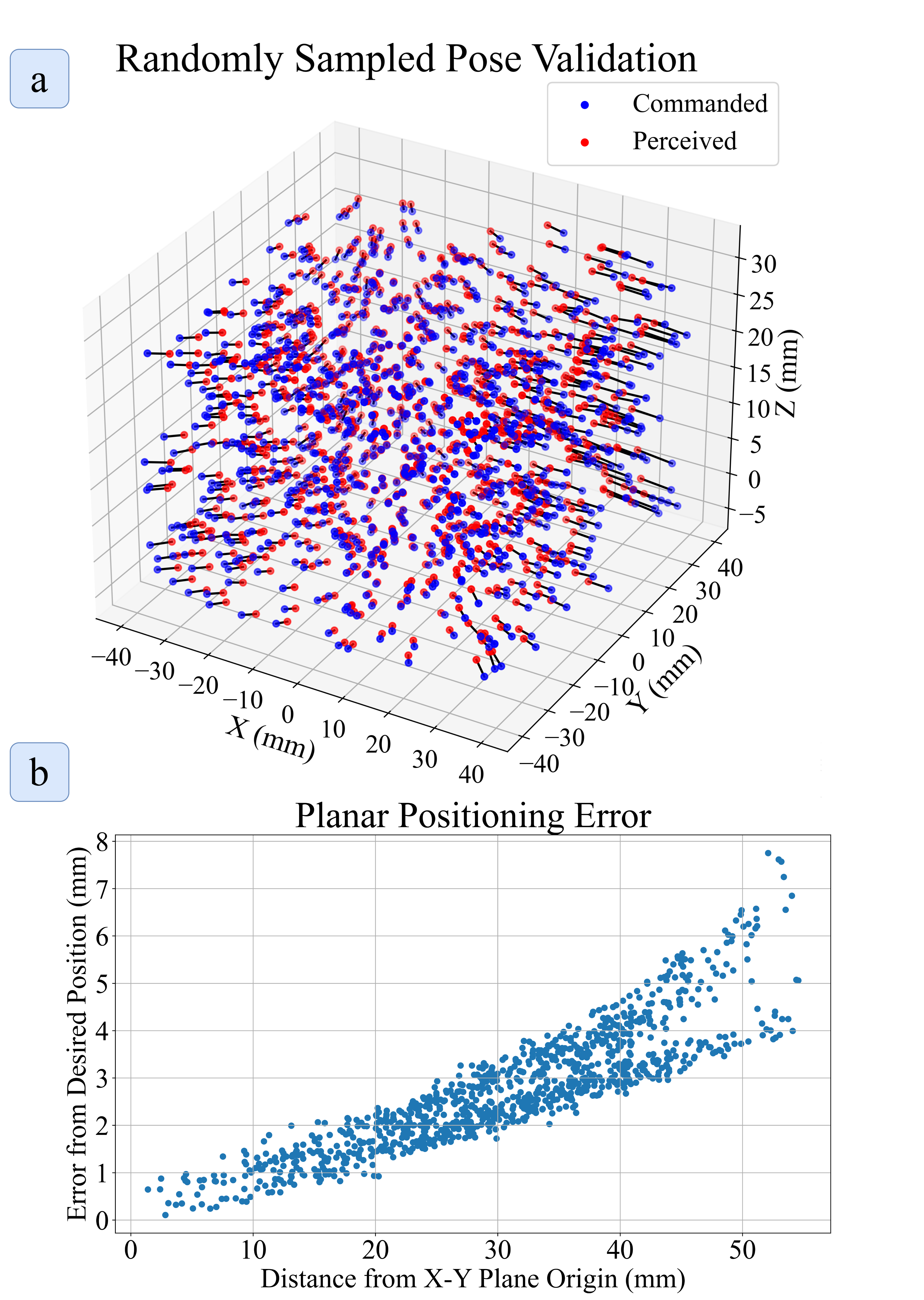}
    \caption{Overview of HASHI motion capture and validation. a) Positioning errors, the line joining pairs of red and blue points indicates the magnitude of the error. b) Spread of errors in the $XY$ plane compared to the distance from the $Z$ axis}
    \label{fig:kinematics_val}
     \vspace{-1.5em}
\end{figure}

\begin{figure*}[t!]
    \vspace{0.5em}
    \includegraphics[width=\linewidth]{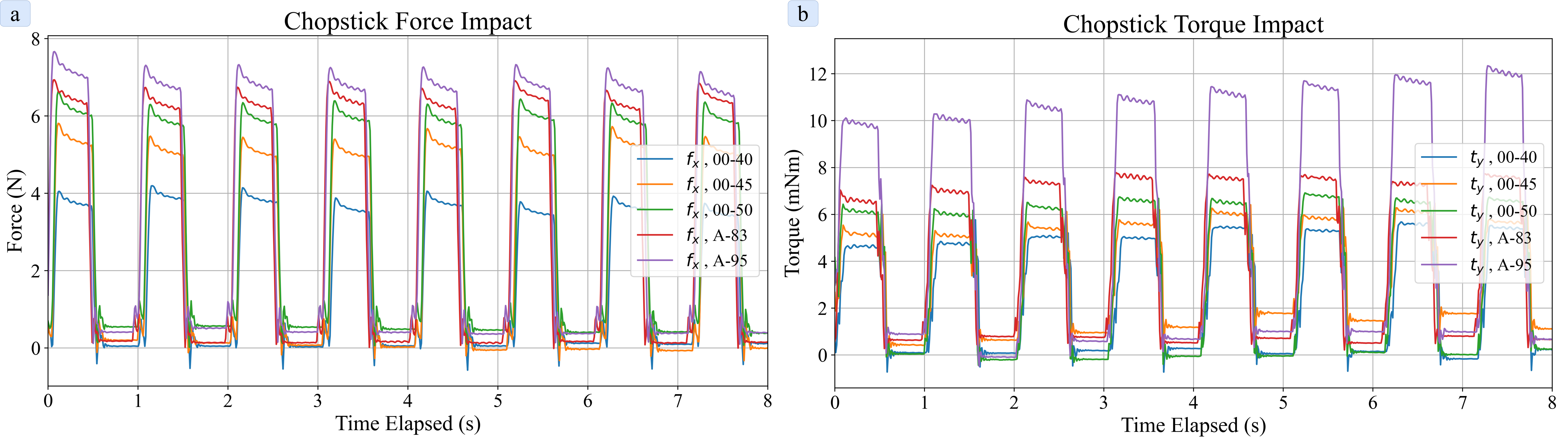}
    \caption{a) X-axis force data for chopstick surface strike against objects of varying shore hardness. b) Observed time-synchronized torque reading from strike.}
    \label{fig:ft_val}
     \vspace{-1.5em}
\end{figure*}

To validate the positional accuracy of HASHI, we fit three retroreflective markers (Qualisys) to a chopstick on a single actuator. The marker representing the tracked position was threaded directly onto the end of the chopstick and then offset appropriately. For the positions of the chopstick, we observed the $\{x,y,z\}$ locations of the tip relative to the zero position of the tip. The orientation of the tip was not considered as the stick orientation is not considered in the inverse kinematic derivation, nor is it relevant to the desired motion. 

We randomly sampled 1000 positions from a uniform distribution in the theoretical workspace of the end effector. This workspace spans $\pm40 \text{mm}$ along the X and Y axes, and $35 \text{mm}$ along the Z axis. Marker positions were recorded using a motion capture system (OptiTrack) and correlated through ROS. The commanded and observed poses are shown in Fig.~\ref{fig:kinematics_val}a. In this figure, a black line connects pairs of correlated positions. The pose error is defined as the $\mathcal{L}_2$ norm between the desired and observed position, given as:
\begin{equation*}
    d = \sqrt{(x_c-x_o)^2+(y_c-y_o)^2+(z_c-z_o)^2}
\end{equation*}
Where, the subscript $c$ is the commanded position in the axis, and the subscript $o$ is the observed position. 

When the chopstick is close to the Z-axis, the error is minimal, as demonstrated in Fig.~\ref{fig:kinematics_val}b. As the tip moves farther in the $XY$ plane away from the origin, the error follows a linearly increasing trend. This is likely due to the backlash of the servo motors and the slop in the linkage mechanism used to drive the backend. Nonetheless, the overall distribution of errors along all axes is minimal for objects of target size ($\leq 20$ mm). The mean error is less than $3$ mm across the workspace (Table \ref{tab:dynamic_lin_vel}), and is smaller toward the center, where two chopsticks make contact with grasped objects. This minimized error in the grasping region is desirable, since it implies that reliable, firm grasping positions are achievable when the device is closed. The larger errors at the edge of the workspace are considered acceptable, as larger strokes of the mechanism at the far extents of its workspace will solely be used for pushing and orienting movements, rather than forming a precise pincer grasp.

From the sampled workspace points as perceived by the motion capture system, we computed a 3D convex hull of the workspace. This result is shown in Fig.~\ref{fig:hashi_mechanism}a to show the reachable positions using the dual chopstick manipulator.

\begin{table}[t]
\vspace{1.0em}
\caption{Kinematic Validation for Individual Manipulator}
\label{tab:dynamic_lin_vel}
\centering
\footnotesize
\setlength\tabcolsep{4 pt} 
\begin{tabular}{c|ccc}
\toprule
    Mean (mm) & x(mm) & y(mm) & z(mm)\\
    \midrule
    $2.93 \pm 1.30$ & $1.88 \pm 1.10$ & $1.79 \pm 1.15$ & $0.79 \pm 0.61$\\
    \bottomrule
    \end{tabular}
    \vspace{-1.0em}
\end{table}

\subsection{Force Feedback Validation}
To verify the sensitivity of the system when grasping deformable objects, we performed a series of grasping tests of objects of varying shore hardness. Platinum cure silicone rubbers (Smooth-On) were cast in a rectangular shape (30 mm $\times$ 40 mm) in rated shore values of 00-40, 00-45, 00-50, A-83, and A-95. Since these objects lie on two different shore scales, where there is no direct meaning in the numerical difference between them, we simply expect to see an increase in force response for firmer objects. Each object was positioned directly between the sticks of the gripper, while they were commanded to close to the zero position. The object was then released, and this routine was repeated for a total of 8 gripping cycles. Fig.~\ref{fig:ft_val} shows the x-axis force and y-axis torque values time-synchronized.

In both the force and torque readings, there is a clear distinction between materials of different shore hardness. Given the spanned range from soft (00-40) to firm (A-95) there is a noticeable increase in observed force along the applied axis as the chopstick makes contact with the test object. These force increases are most pronounced in the softer materials (00-40, 00-45, 00-50). The difference between soft objects ($\approx 2.25$ N) is more pronounced than the firmer shore A rated objects ($\approx 0.8$ N). Significantly, these results show that HASHI has excellent sensitivity to distinguish between objects with small differences in hardness. For applications such as manipulation of ripe or unripe fruit, this ability to perceive hardness may be consequential to the chosen manipulation strategy (e.g. pinch vs. lift under).
\section{Grasping Experiments}
\label{sec:exper}
We evaluate the performance of the HASHI gripper by designing a set of experiments inspired by the YCB grasping benchmark \cite{bekiroglu2019benchmarking}; however, we redesign the set of objects to consist of small, deformable food items. Instead of using a grasp planning algorithm, the robot and gripper are teleoperated into a repeatable pregrasp pose. Items are weighed and measured before being placed on a small raised platform. While positioned parallel to the table, the chopsticks are closed around the target food item, and then programmatically held to this final closure position. The items are then lifted 25 cm above the table and linearly translated 20 cm at $0.2$m/s for three cycles. For the rotation tests, the food is lifted in the same way, but at the apex of the motion upwards, the end effector is instead rotated $90^{\circ}$ about the $Y$ and $Z$ axes. If the object remains in the gripper during translation and rotation, the subcomponents are marked as success (Y). If the object slips out of the gripper or disintegrates at any point during the motion, the trial is marked as failure (N). Table \ref{tab:results_robot} shows the full results of the grasping experiments and the parameters for the items.
\begin{table}
    \centering
    \small
    \begin{tabular}{|| c | c | c | c | c ||}
        \hline
        Food & Weight (g) & Dim.(mm) & Rot.&  Lin.\\
        \hline \hline
        Sushi Roll & 25 & 35$\times$45$\times$25 & \cellcolor{green!25}Y& \cellcolor{green!25}Y\\
        \hline
        Frozen Scallop & 30 & 50$\times$45$\times$20 & \cellcolor{red!25}N& \cellcolor{red!25}N\\
        \hline
        Raw Scallop & 38 & 50$\times$45$\times$20 & \cellcolor{green!25}Y& \cellcolor{green!25}Y\\
        \hline
        Cooked Scallop & 25 & 40$\times$35$\times$25 & \cellcolor{green!25}Y& \cellcolor{green!25}Y\\
        \hline
        Grape & 9 & 28$\times$18$\times$24 & \cellcolor{green!25}Y& \cellcolor{green!25}Y\\
        \hline
        Edamame & $<1$ & 15$\times$10$\times$8 & \cellcolor{green!25}Y& \cellcolor{green!25}Y\\
        \hline
        Potsticker & 29 & 90$\times$27$\times$40 & \cellcolor{green!25}Y& \cellcolor{green!25}Y\\
        \hline
        Salmon Sashimi & 13 & 90$\times$35$\times$5 & \cellcolor{green!25}Y& \cellcolor{green!25}Y\\
        \hline
        Tuna Nigiri & 50 & 90$\times$35$\times$40 & \cellcolor{red!25}N& \cellcolor{red!25}N\\
        \hline
        Broccoli & 12 & 60$\times$50$\times$40 & \cellcolor{green!25}Y& \cellcolor{green!25}Y\\
        \hline
        Shrimp (small) & 7 & 48$\times$35$\times$14 & \cellcolor{red!25}N& \cellcolor{red!25}N\\
        \hline
        Shrimp (med) & 14 & 65$\times$45$\times$15 & \cellcolor{red!25}N& \cellcolor{green!25}Y\\
        \hline
    \end{tabular}
    \normalsize
    \caption{Food parameters and results of grasping trials.}
    \label{tab:results_robot}
    \vspace{-1.5em}
\end{table}

HASHI performed well during the majority of the trials. The nigiri trials were the only ones where the food item slipped out during movement, first because the grasp wasn't in the center of the heavy rice base, and second because the tuna on top wasn't secured. We hypothesize that transport of an object like this is possible, provided it is grasped at an angle or orthogonal to the table, similar to how it would be grasped in reality. The frozen scallop also proved to be difficult, namely due to its lubricity and rigidity. Round or spherical objects normally considered to be unstable when grasped at two points were quite secure during trials, likely due to the slight deformability we were able to exploit. Some additional foodstuffs were qualitatively manipulated as shown in Fig. \ref{fig:HASHI_first_page}.

\section{Conclusion}
\label{sec:conc}

In this work, we demonstrated the design, modeling, and validation of a chopstick-inspired dexterous end effector. Our mechanical design enables the manipulation of a wide variety of food items while also maintaining excellent force perception near the grasping interface. As the kinematics are based on individual chopsticks, this two-stick design is extensible into configurations that utilize multiple sticks for enhanced stability and in-hand manipulation capability. Future work will also include the incorporation of HASHI into grasp planning mechanisms for motion planning, and the development of other manipulation strategies, such as pushing, scooping, and piercing. Additionally, the potential of F/T feedback in the servo control loop will offer future opportunities for precision grasping and enhanced dexterity. In sum, HASHI represents an exciting innovation in end effector design by combining mechanical simplicity with rich dexterous control. 
\section*{Acknowledgements}
The authors are grateful to Rui Luo for his insights on the kinematics derivation. Adriana Cespedes, and Vedant Rautela for their assistance in the kinematic validation procedures. Ethan Holand kindly advised on the rendering of the 3D models.
\bibliographystyle{IEEEtran} 
\bibliography{references}

\begin{thebibliography}{10}
\providecommand{\url}[1]{#1}
\csname url@samestyle\endcsname
\providecommand{\newblock}{\relax}
\providecommand{\bibinfo}[2]{#2}
\providecommand{\BIBentrySTDinterwordspacing}{\spaceskip=0pt\relax}
\providecommand{\BIBentryALTinterwordstretchfactor}{4}
\providecommand{\BIBentryALTinterwordspacing}{\spaceskip=\fontdimen2\font plus
\BIBentryALTinterwordstretchfactor\fontdimen3\font minus
  \fontdimen4\font\relax}
\providecommand{\BIBforeignlanguage}[2]{{%
\expandafter\ifx\csname l@#1\endcsname\relax
\typeout{** WARNING: IEEEtran.bst: No hyphenation pattern has been}%
\typeout{** loaded for the language `#1'. Using the pattern for}%
\typeout{** the default language instead.}%
\else
\language=\csname l@#1\endcsname
\fi
#2}}
\providecommand{\BIBdecl}{\relax}
\BIBdecl

\bibitem{gephart2019create}
J.~A. Gephart, H.~E. Froehlich, and T.~A. Branch, ``To create sustainable
  seafood industries, the united states needs a better accounting of imports
  and exports,'' \emph{Proceedings of the National Academy of Sciences}, vol.
  116, no.~19, pp. 9142--9146, 2019.

\bibitem{andrychowicz2020learning}
O.~M. Andrychowicz, B.~Baker, M.~Chociej, R.~Jozefowicz, B.~McGrew,
  J.~Pachocki, A.~Petron, M.~Plappert, G.~Powell, A.~Ray \emph{et~al.},
  ``Learning dexterous in-hand manipulation,'' \emph{The International Journal
  of Robotics Research}, vol.~39, no.~1, pp. 3--20, 2020.

\bibitem{Cai2023}
Y.~Cai and S.~Yuan, ``In-hand manipulation in power grasp: Design of an
  adaptive robot hand with active surfaces,'' in \emph{2023 IEEE International
  Conference on Robotics and Automation (ICRA)}, 2023, pp. 10\,296--10\,302.

\bibitem{wang2017prestressed}
Z.~Wang, Y.~Torigoe, and S.~Hirai, ``A prestressed soft gripper: design,
  modeling, fabrication, and tests for food handling,'' \emph{IEEE Robotics and
  Automation Letters}, vol.~2, no.~4, pp. 1909--1916, 2017.

\bibitem{Ishikawa2022}
R.~Ishikawa, M.~Hamaya, F.~Von~Drigalski, K.~Tanaka, and A.~Hashimoto,
  ``Learning by breaking: Food fracture anticipation for robotic food
  manipulation,'' \emph{IEEE Access}, vol.~10, pp. 99\,321--99\,329, 2022.

\bibitem{Williams2005}
\BIBentryALTinterwordspacing
S.~H. Williams, B.~W. Wright, V.~d. Truong, C.~R. Daubert, and C.~J. Vinyard,
  ``Mechanical properties of foods used in experimental studies of primate
  masticatory function,'' \emph{American Journal of Primatology}, vol.~67,
  no.~3, pp. 329--346, 2005. [Online]. Available:
  \url{https://onlinelibrary.wiley.com/doi/abs/10.1002/ajp.20189}
\BIBentrySTDinterwordspacing

\bibitem{kadowaki2013}
R.~Kadowaki, N.~Inou, H.~Kimura \emph{et~al.}, ``Measurement of microscopic
  young's modulus of crispy foods.'' \emph{International Proceedings of
  Chemical, Biological and Environmental Engineering (IPCBEE)}, vol.~50, pp.
  79--83, 2013.

\bibitem{Bhattacharjee2019}
T.~Bhattacharjee, G.~Lee, H.~Song, and S.~S. Srinivasa, ``Towards robotic
  feeding: Role of haptics in fork-based food manipulation,'' \emph{IEEE
  Robotics and Automation Letters}, vol.~4, no.~2, pp. 1485--1492, 2019.

\bibitem{park2016towards}
D.~Park, Y.~K. Kim, Z.~M. Erickson, and C.~C. Kemp, ``Towards assistive feeding
  with a general-purpose mobile manipulator,'' \emph{arXiv preprint
  arXiv:1605.07996}, 2016.

\bibitem{Wang2022}
\BIBentryALTinterwordspacing
Z.~Wang, S.~Hirai, and S.~Kawamura, ``Challenges and opportunities in robotic
  food handling: A review,'' \emph{Frontiers in Robotics and AI}, vol.~8, 2022.
  [Online]. Available:
  \url{https://www.frontiersin.org/articles/10.3389/frobt.2021.789107}
\BIBentrySTDinterwordspacing

\bibitem{Qiu2023}
A.~Qiu, C.~Young, A.~L. Gunderman, M.~Azizkhani, Y.~Chen, and A.-P. Hu,
  ``Tendon-driven soft robotic gripper with integrated ripeness sensing for
  blackberry harvesting,'' in \emph{2023 IEEE International Conference on
  Robotics and Automation (ICRA)}, 2023, pp. 11\,831--11\,837.

\bibitem{Low2022}
J.~H. Low, P.~M. Khin, Q.~Q. Han, H.~Yao, Y.~S. Teoh, Y.~Zeng, S.~Li, J.~Liu,
  Z.~Liu, P.~Valdivia~y Alvarado, I.-M. Chen, B.~C.~K. Tee, and R.~C.~H. Yeow,
  ``Sensorized reconfigurable soft robotic gripper system for automated food
  handling,'' \emph{IEEE/ASME Transactions on Mechatronics}, vol.~27, no.~5,
  pp. 3232--3243, 2022.

\bibitem{ZWang2021}
\BIBentryALTinterwordspacing
Z.~Wang, H.~Furuta, S.~Hirai, and S.~Kawamura, ``A scooping-binding robotic
  gripper for handling various food products,'' \emph{Frontiers in Robotics and
  AI}, vol.~8, 2021. [Online]. Available:
  \url{https://www.frontiersin.org/articles/10.3389/frobt.2021.640805}
\BIBentrySTDinterwordspacing

\bibitem{Amend2016}
\BIBentryALTinterwordspacing
J.~Amend, N.~Cheng, S.~Fakhouri, and B.~Culley, ``Soft robotics
  commercialization: Jamming grippers from research to product,'' \emph{Soft
  Robotics}, vol.~3, no.~4, pp. 213--222, 2016, pMID: 28078197. [Online].
  Available: \url{https://doi.org/10.1089/soro.2016.0021}
\BIBentrySTDinterwordspacing

\bibitem{Gafer2020}
A.~Gafer, D.~Heymans, D.~Prattichizzo, and G.~Salvietti, ``The quad-spatula
  gripper: A novel soft-rigid gripper for food handling,'' in \emph{2020 3rd
  IEEE International Conference on Soft Robotics (RoboSoft)}, 2020, pp. 39--45.

\bibitem{Petterson2010}
A.~Petterson, T.~Ohlsson, D.~G. Caldwell, S.~Davis, J.~O. Gray, and T.~J. Dodd,
  ``A bernoulli principle gripper for handling of planar and 3d (food)
  products,'' \emph{Industrial robot an international journal.}, vol.~37,
  no.~6, 2010-10-19.

\bibitem{Wang2021}
Z.~Wang, Y.~Makiyama, and S.~Hirai, ``A soft needle gripper capable of grasping
  and piercing for handling food materials,'' \emph{Journal of Robotics and
  Mechatronics}, vol.~33, no.~4, pp. 935--943, 2021.

\bibitem{Endo2016}
G.~Endo and N.~Otomo, ``Development of a food handling gripper considering an
  appetizing presentation,'' in \emph{2016 IEEE International Conference on
  Robotics and Automation (ICRA)}, 2016, pp. 4901--4906.

\bibitem{Yang2022}
\BIBentryALTinterwordspacing
Z.~Yang, K.~Yin, and L.~Liu, ``Learning to use chopsticks in diverse gripping
  styles,'' \emph{ACM Trans. Graph.}, vol.~41, no.~4, jul 2022. [Online].
  Available: \url{https://doi.org/10.1145/3528223.3530057}
\BIBentrySTDinterwordspacing

\bibitem{Marcosticks2021}
\BIBentryALTinterwordspacing
Staff, ``Ten thousand ways to use chopsticks,'' Jun 2021. [Online]. Available:
  \url{https://marcosticks.org/poster-ten-thousand-ways-to-use-chopsticks/}
\BIBentrySTDinterwordspacing

\bibitem{Ke2020}
L.~Ke, A.~Kamat, J.~Wang, T.~Bhattacharjee, C.~Mavrogiannis, and S.~S.
  Srinivasa, ``Telemanipulation with chopsticks: Analyzing human factors in
  user demonstrations,'' in \emph{2020 IEEE/RSJ International Conference on
  Intelligent Robots and Systems (IROS)}, 2020, pp. 11\,539--11\,546.

\bibitem{Ke2021}
L.~Ke, J.~Wang, T.~Bhattacharjee, B.~Boots, and S.~Srinivasa, ``Grasping with
  chopsticks: Combating covariate shift in model-free imitation learning for
  fine manipulation,'' in \emph{2021 IEEE International Conference on Robotics
  and Automation (ICRA)}, 2021, pp. 6185--6191.

\bibitem{Dextrous}
\BIBentryALTinterwordspacing
 [Online]. Available: \url{https://www.dextrousrobotics.com/robots}
\BIBentrySTDinterwordspacing

\bibitem{Tanikawa1999}
T.~Tanikawa and T.~Arai, ``Development of a micro-manipulation system having a
  two-fingered micro-hand,'' \emph{IEEE Transactions on Robotics and
  Automation}, vol.~15, no.~1, pp. 152--162, 1999.

\bibitem{Yabugaki2013}
H.~Yabugaki, K.~Ohara, M.~Kojima, Y.~Mae, T.~Tanikawa, and T.~Arai, ``Automated
  stable grasping with two-fingered microhand using micro force sensor,'' in
  \emph{2013 IEEE International Conference on Robotics and Automation}, 2013,
  pp. 2771--2776.

\bibitem{Sakurai2016}
H.~Sakurai, T.~Kanno, and K.~Kawashima, ``Thin-diameter chopsticks robot for
  laparoscopic surgery,'' in \emph{2016 IEEE International Conference on
  Robotics and Automation (ICRA)}, 2016, pp. 4122--4127.

\bibitem{Tadano2010}
K.~Tadano, K.~Kawashima, K.~Kojima, and N.~Tanaka, ``Development of a pneumatic
  surgical manipulator ibis iv,'' \emph{Journal of Robotics and Mechatronics},
  vol.~22, no.~2, p. 179, 2010.

\bibitem{koshizaki2010control}
T.~Koshizaki and R.~Masuda, ``Control of a meal assistance robot capable of
  using chopsticks,'' in \emph{ISR 2010 (41st International Symposium on
  Robotics) and ROBOTIK 2010 (6th German Conference on Robotics)}.\hskip 1em
  plus 0.5em minus 0.4em\relax VDE, 2010, pp. 1--6.

\bibitem{yamazaki2012autonomous}
A.~Yamazaki and R.~Masuda, ``Autonomous foods handling by chopsticks for meal
  assistant robot,'' in \emph{ROBOTIK 2012; 7th German Conference on
  Robotics}.\hskip 1em plus 0.5em minus 0.4em\relax VDE, 2012, pp. 1--6.

\bibitem{yamazaki2012various}
A.~"Yamazaki and R.~Masuda, ``Various foods handling movement of
  chopstick-equipped meal assistant robot and there evaluation,'' in
  \emph{Social Robotics}, S.~S. Ge, O.~Khatib, J.-J. Cabibihan, R.~Simmons, and
  M.-A. Williams, Eds.\hskip 1em plus 0.5em minus 0.4em\relax Berlin,
  Heidelberg: Springer Berlin Heidelberg, 2012, pp. 158--167.

\bibitem{quigley2009ros}
M.~Quigley, K.~Conley, B.~Gerkey, J.~Faust, T.~Foote, J.~Leibs, R.~Wheeler,
  A.~Y. Ng \emph{et~al.}, ``Ros: an open-source robot operating system,'' in
  \emph{ICRA workshop on open source software}, vol.~3, no. 3.2.\hskip 1em plus
  0.5em minus 0.4em\relax Kobe, Japan, 2009, p.~5.

\bibitem{denavit1955kinematic}
J.~Denavit and R.~S. Hartenberg, ``A kinematic notation for lower-pair
  mechanisms based on matrices,'' 1955.

\bibitem{Wolfram}
\BIBentryALTinterwordspacing
E.~W. Weisstein, ``Spherical coordinates. {From MathWorld---A Wolfram Web
  Resource},'' last visited on 15/09/2023. [Online]. Available:
  \url{https://mathworld.wolfram.com/SphericalCoordinates.html}
\BIBentrySTDinterwordspacing

\bibitem{bekiroglu2019benchmarking}
Y.~Bekiroglu, N.~Marturi, M.~A. Roa, K.~J.~M. Adjigble, T.~Pardi, C.~Grimm,
  R.~Balasubramanian, K.~Hang, and R.~Stolkin, ``Benchmarking protocol for
  grasp planning algorithms,'' \emph{IEEE Robotics and Automation Letters},
  vol.~5, no.~2, pp. 315--322, 2019.

\end{thebibliography}

\addtolength{\textheight}{-12cm}   




\end{document}